# LOW COMPLEXITY CONVOLUTIONAL NEURAL NETWORK FOR VESSEL SEGMENTATION IN PORTABLE RETINAL DIAGNOSTIC DEVICES

*M. Hajabdollahi[1], R. Esfandiarpoor[1], S.M.R. Soroushmehr[2], N. Karimi[1], S. Samavi[1, 3], K. Najarian[2, 3]*

[1]Department of Electrical and Computer Engineering, Isfahan University of Technology, Isfahan 84156-83111, Iran
[2]Department of Computational Medicine and Bioinformatics, University of Michigan, Ann Arbor, U.S.A.
[3]Michigan Center for Integrative Research in Critical Care, University of Michigan, Ann Arbor, U.S.A.

## ABSTRACT

Retinal vessel information is helpful in retinal disease screening and diagnosis. Retinal vessel segmentation provides useful information about vessels and can be used by physicians during intraocular surgery and retinal diagnostic operations. Convolutional neural networks (CNNs) are powerful tools for classification and segmentation of medical images. Complexity of CNNs makes it difficult to implement them in portable devices such as binocular indirect ophthalmoscopes. In this paper a simplification approach is proposed for CNNs based on combination of quantization and pruning. Fully connected layers are quantized and convolutional layers are pruned to have a simple and efficient network structure. Experiments on images of the STARE dataset show that our simplified network is able to segment retinal vessels with acceptable accuracy and low complexity.

## 1. INTRODUCTION

Most of the eye diseases and loss of vision are due to the diabetes [1]. Analysis and screening of the retinal vessels are very useful in eye disease detection and diagnosis. Diabetic retinopathy (DR) can be controlled with regular examination of eyes in early stages [1]. By the increase of DR, automatic detection of vessel has become important in telemedicine applications. Having proper knowledge about retinal vessels also can be helpful during any retinal surgery operations [2]. As illustrated in Fig. 1 automatic segmentation of vessels can be helpful during intraocular surgery [3].

The problem of retinal vessel segmentation is investigated by many researchers in recent years. In [4] local thresholding using gray-scale level intensity as well as spatial dependencies is used for coarse segmentation of the vessel points. In [5] a match filter based on the wavelet kernels is proposed to identify the vessel borders. In [6] image blocks are summarized as vectors in each direction and a match filter is utilized to provide seeds for region growing algorithm. In [7] a combination of shifted and blurred DOG filters is used to detect the vessel points.

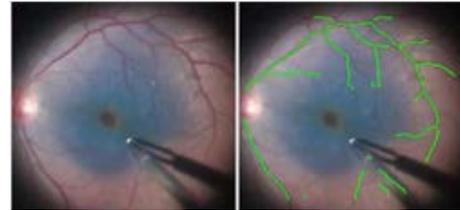

Fig. 1. Retinal vessel segmentation during surgery [3].

Convolutional neural networks have been introduced as a powerful tool for medical image analysis. In this regard [8,9,10,11] are focused on the problem of vessel segmentation using CNNs. In [8] CNN is used for segmentation of a whole patch. In [9] CNN is trained on a big dataset containing 400,000 retinal image samples. In [10] probability map generated from CNN is enhanced using conditional random field (CRF). In [11] a multi-level CNN is utilized to represent the feature hierarchies. CNN layers are equipped with a side output layer to approximate ground truth by minimizing a loss function.

Vessel segmentation in intraocular surgery requires real time vessel analysis and for portable and onsite devices needs real time and low power segmentation system [3, 12, 13]. In this regard some recent studies focused on the problem of real time retinal vessel segmentation on dedicated hardware such as GPU and FPGA. In [14, 15] methods for retinal vessel analysis are presented and implemented on GPU. In [12] and [13] hardware architectures are presented for retinal vessel segmentation system.

Although CNN has been introduced as state of the art methods in classification problems but its structure is complex and requires a lot of arithmetic operations. Simplification methods recently have been developed on the CNN structure. In [16] a framework for fixed point representation of CNN, and in [17] a method for pruning connections, are proposed. Also binarized neural networks are introduced as another way to overcome the complexity of CNNs [18, 19]. Applying CNN in portable devices for retinal vessel analysis such as binocular indirect ophthalmoscope requires overcoming the complexity problem of CNN.

In this paper, problem of simplifying CNN as a powerful classifier for retinal image segmentation is addressed. The proposed method is based on two techniques including pruning and quantization. Fully connected layers are quantized and convolutional layers are pruned effectively. The reminder of this paper is organized as follows. In Section 2, proposed method for retinal vessel segmentation is proposed. Section 3 is dedicated to the experimental results. Finally in Section 4 the concluding remarks are presented.

## 2. PROPOSED METHOD

In Fig. 2 overview of the system for segmentation of retinal blood vessels is presented. A convolutional neural network (CNN) structure is utilized for vessel segmentation in enhanced retinal images. As it is observed in Fig. 2 at first fully connected layers are modified to have low complexity. After that, convolutional layers are pruned and unnecessary weights are removed. Proposed method stages are described in more details as follows.

### 2.1. Retinal Image Enhancement

Low contrast retinal images can be enhanced using histogram equalization as preprocessing [7, 1]. Green channel was considered as the most representative channel in RGB fundus images [1]. In Fig. 3 (a) a sample retinal image in form of RGB, R, G and B are represented. In Fig. 3 (b) images of Fig. 3 (a) are enhanced using local histogram equalization. Considering enhanced RGB image as input of CNN, causes a dramatic increase in the complexity of the network structure. To have the best and enough information as well as simple network structure, enhanced gray-scale level of image is selected. Enhanced gray-scale level image which has average information of the all channels is considered as input of the CNN for vessel segmentation.

### 2.2. Simplified CNN

Main parts of a typical CNN are convolutional layers (CLs) and fully convolutional layers (FCLs). CLs consume the most of the arithmetic operation parts while FCLs have the majority of the parameters [16]. These two parts must be taken into consideration for simplification of the CNN structure. Recently quantization and pruning were used as two ways for reducing the structural complexity of the neural networks. In the proposed method a hybrid method for simplification of CNN including quantization and pruning is presented. In the following subsections proposed method for quantization and punning is presented.

#### 2.2.1. Quantization

In binarization, a value converts to two possible values such as 0 and 1. In [18, 19] deterministic and stochastic methods are proposed for weight binarization as (1) and (2).

$$W_b = \begin{cases} +1 & if\ W \geq 0, \\ -1 & otherwise. \end{cases} \quad (1)$$

$$W_b = \begin{cases} +1 & with\ probability\ p = \sigma(w), \\ -1 & with\ probability\ 1 - p. \end{cases} \quad (2)$$

$$\sigma(x) = clip\left(\frac{x+1}{2}, 0, 1\right) = \max(0, \min(1, \frac{x+1}{2})) \quad (3)$$

In (1) and (2), W and $W_b$ are original and binarized network weights respectively. In (3), σ (x) is the "hard sigmoid function". In this paper deterministic ternary quantization of the weights is applied as (4).

$$W_b = \begin{cases} -1 & if\ W < 0 \\ 0 & if\ W = 0 \\ 1 & if\ W > 0 \end{cases} \quad (4)$$

As it is observed from Fig. 2 after CNN training, only FCLs are quantized in form of (4). Quantization without retraining causes a significant drop of accuracy. Hence after quantization, CNN is retrained. The process of quantization-retraining is repeated while suitable accuracy is obtained. 32-bit representation consumes 32 times more memory size and memory accesses than the binary representation [20]. Quantized representation with 2 bits significantly reduces energy consumption of the arithmetic operations [20].

Although quantization significantly reduces the network complexity, but quantization of convolutional layers causes degradation of the network's learning capability. Hence in the proposed method, quantization is performed only on weights of the fully connected layers. Still the convolutional layers would remain complex but all multiplications in FCLs are replaced with simple transfer operations.

#### 2.2.2. Pruning

Pruning is yet another way of reducing the network's complexity which can eliminate un-necessary weights during the network's training [17]. Basic pruning schema has three steps. At first, network weights and connections are trained. Weight connections smaller than a threshold value are eliminated from the network. Finally the loss of accuracy due to weigh eliminations is recovered by retraining. This process is continued until the network come to a stable station.

CL quantization has undesirable effect on network performance however pruning removes the un-necessary CL connections. Pruning CL connections can be introduced as an efficient way to overcome the problem of large number

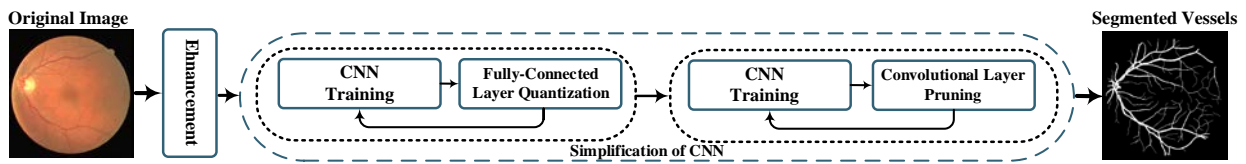

Fig. 2. Proposed simplified CNN for retinal vessel segmentation.

of arithmetic operations in the convolutional layers.

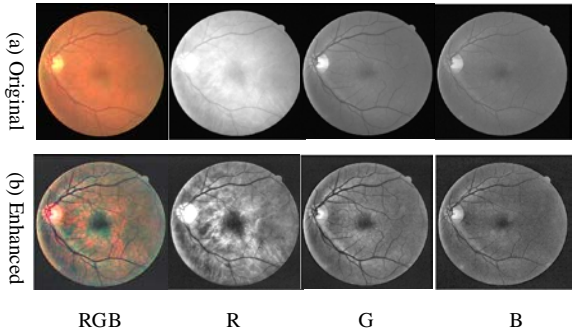

Fig. 3. Local histogram equalization

In Fig. 4 pruning in CL level is illustrated. Using the basic pruning schema [17], CLs are pruned after FCL quantization. As illustrated in Fig. 4 in CL pruning step, all weights smaller than a threshold value are eliminated. This threshold is obtained by production of a constant value in variance of the weights in each convolutional layer. After each pruning step, network is retrained to restore the effect of the eliminated weights. Retraining is performed in both FCL and CL weights by preserving FCL weight quantization.

Due to the operation of the convolutional filters on the entire of an image, pruning the convolutional filters is very useful to reduce the complexity of the network structure. In the proposed hybrid method, all parameters of FCLs are converted to 0, 1 or -1. After quantization, CLs complexity is further reduced by pruning.

During quantization, based on (4), all of the weights are changed regardless of their values, while the pruning procedure only removes connections with low weight values. With binarized network one bit operations are required which is useful for the simplicity of the network. But, drastic changes of the weights caused by quantization may have destructive effect on the network learning capability. Although bit length parameters are long in the pruned network, but pruning creates lower changes than the quantization. A combination of these two methods including quantization and pruning can be considered as an effective way to simplify the network structure.

## 3. EXPERIMENTAL RESULTS

For evaluation of the proposed method, experimental results are performed using TensorFlow framework. Simplified CNN is trained and tested on the STARE image dataset with the first observer segmentation [21]. Image patches with size 9×9 from the enhanced gray-scale level images are extracted around each pixel and make the network input. Output of the each patch is set as class of the pixel under consideration and CNN is trained as a binary classification problem. 5-fold cross validation method is used for performance validation. Accuracy, sensitivity, specificity and ROC are used for classification performance evaluation. Experimental results are organized in three parts as follows.

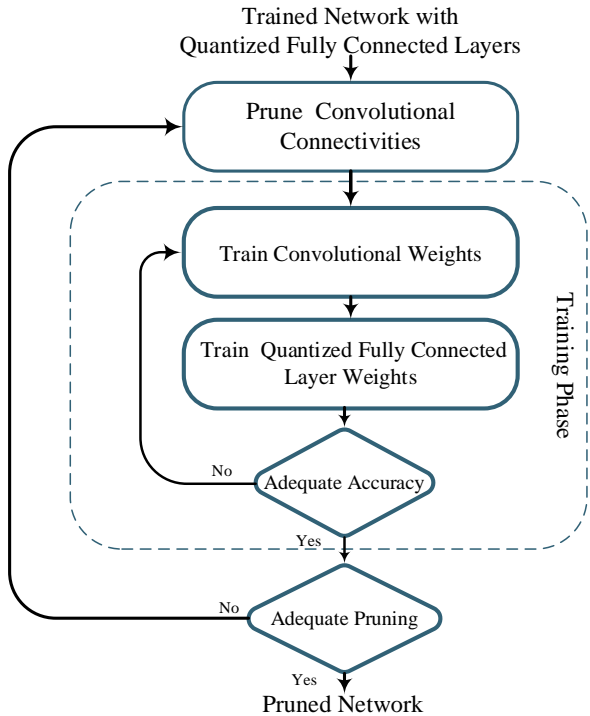

Fig. 4. Pruning of the convolutional layers.

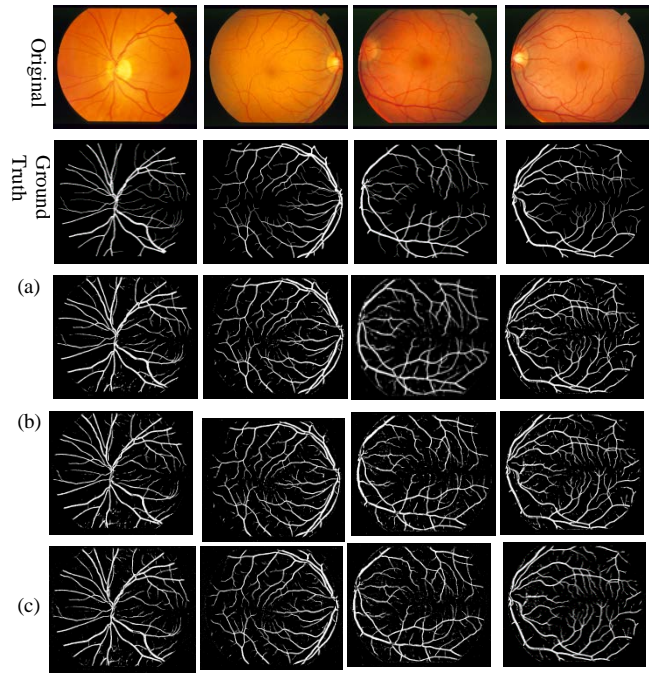

Fig. 5. Visual illustration of segmnetaiotn. (a) CNN with original parameters, (b) CNN with FCLs quantization, (c) CNN with FCLs quantization and CLs pruning.

## 3.1. CNN with original parameters

Different CNN configurations are experimented with low structure complexity consideration. Finally a CNN with low complexity is applied. Visual results in case of some image samples from STARE [21] are illustrated in Fig. 5 (a). Performance results of CNN without any simplification (original parameters) are compared with related works in Table 1. As illustrated in Table 1 the network has suitable parameters without any simplification. DICE score and accuracy are achieved to be about 0.76 and 0.96, respectively. In Fig. 6 receiver operating characteristic (ROC) of the classification in CNN with original parameters is depicted. Although the performance results is desirable but the network structure has not been simplified. Applying the segmentation method on the portable devices with the limited hardware resources requires a simple and fast network structure.

## 3.2. CNN Quantization

CNN after training is quantized using three levels including 0,1 and -1 as mentioned in (4). Firstly, both convolutional layers and fully connected layers are quantized as (4). As it was mentioned in the Section 2.2, quantization changes the weights drastically and it may lead to the learning disability. It is observed that the fully quantized network cannot reach to a suitable performance and accuracy of about 0.75 is achieved. Therefore only the fully connected layers are quantized. It can be observed from Fig. 5 (b) that CNN with quantized fully connected layers is able to segment the vessel points approximately same as it's original. Visual results are not significantly different in Fig. 5 (a) and Fig. 5 (b). Also performance of the quantized FCLs in Table 1 and Fig. 6 is comparable with CNN with original parameters. For example accuracy reduction of 0.003 is resulted as compared with the original CNN.

## 3.3. CNN training with CLs pruning and FCLs quantization

Although the FCLs are quantized but the complexity of the network in CLs is still high. Therefore after quantization,

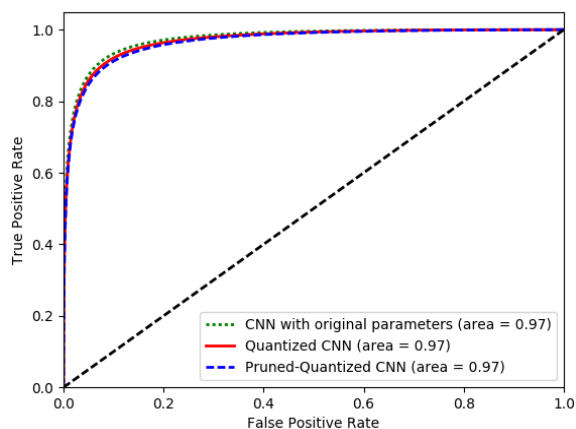

Fig. 6. ROC curve of experimented CNNs

Table 1. Performance comparison with related works on STARE

| Method | SEN | SPE | ACC |
|---|---|---|---|
| [7] (2015) | 0.7716 | 0.9701 | 0.9497 |
| [11] (2016) | 0.7412 | -- | 0.9585 |
| [10] (2016) | 0.7140 | -- | 0.9545 |
| [15] (2014) | 0.7305 | 0.9688 | 0.9440 |
| [13] (2018) | 0.7538 | 0.9608 | 0.9440 |
| Proposed (CNN with original parameters) | 0.7823 | 0.9770 | 0.9617 |
| Proposed (Quantized CNN) | 0.7792 | 0.9740 | 0.9587 |
| Proposed (Pruned-Quantized CNN) | 0.7599 | 0.9757 | 0.9581 |

filter weights are pruned as Fig. 4 to eliminate the unnecessary weights for the simpler structure. Visual results in Fig. 5 (c) and ROC result in Fig. 6 represent no significant difference in the performance. Also Table 1 shows 0.0036 accuracy reduction in case of pruned-quantized CNN. Finally in Table 2 network structure is presented before and after simplification. In CLs about 60% of the weights is removed by pruning and quantization causes FCLs weights to have a ternary values.

## 4. CONCLUSION

In this paper convolutional neural network for automatic segmentation of the vessel in fundus images was presented. The structure of the CNN was simplified in such a way to make it suitable for hardware implementation in portable and onsite retinal diagnostic devices. Combination of two simplification mechanisms was applied. Fully connected layers were quantized while convolutional layers were pruned. Simulation results in case of STARE dataset demonstrated that the acceptable performance for simplified CNN was obtained. Finally in CNN with RUC of 0.97, we removed 60% of the convolutional layer weights and all of the fully connected layer weights were quantized. Proposed simplified CNN could be considered as a method for automatic segmentation of vessels in portable retinal diagnostic devices.

Table 2. Structure of original and simplified CNNs.

| Layer | Type | Maps and Neurons | Filter Size | Original Weights | Simplified Weights |
|---|---|---|---|---|---|
| 1 | Input | 1M × 9×9N | - | - | - |
| 2 | Convolution | 64M × 9×9N | 3×3 | 576 | 395 |
| 3 | Max Pooling | 64M × 5×5N | 2×2 | - | - |
| 4 | Convolution | 32M × 5×5N | 3×3 | 18432 | 7555 |
| 5 | Max Pooling | 32M × 3×3N | 2×2 | - | - |
| 6 | FC | 50N | 1×1 | 14400 | Quantized |
| 7 | FC | 20N | 1×1 | 1000 | Quantized |
| 8 | FC | 2N | 1×1 | 40 | Quantized |